\documentclass[letterpaper, 10 pt, conference]{ieeeconf}  % Comment this line out if you need a4paper

\IEEEoverridecommandlockouts                              % This command is only needed if 
\usepackage{graphics} % for pdf, bitmapped graphics files
\usepackage{epsfig} % for postscript graphics files
\usepackage{amsmath} % assumes amsmath package installed
\usepackage{amssymb}  % assumes amsmath package installed
\usepackage{makecell}
\usepackage{multicol}
\usepackage{siunitx}
\usepackage[caption=false,font=footnotesize]{subfig}
\usepackage{hyperref}
\usepackage{graphicx}
\usepackage[nolist,nohyperlinks]{acronym}
\usepackage{siunitx}
\usepackage{verbatim}
\usepackage{todonotes}
\usepackage[noend]{algpseudocode}
\usepackage{algorithm}

\title{\LARGE \bf
Combined Sampling and Optimization Based Planning for Legged-Wheeled Robots
}

\author{Edo Jelavic, Farbod Farshidian, and Marco Hutter% <-this % stops a space
\thanks{This research was supported by the Swiss National Science Foundation (SNSF) as part of project No.188596, by the European Research Council (ERC) under the European Union’s Horizon 2020 research and innovation programme grant agreement No 852044 and through the SNSF National Centre of Competence in Digital Fabrication (NCCR dfab).}% <-this % stops a space
\thanks{All authors are with the Robotic Systems Lab, ETH Zurich, 8092 Zurich, Switzerland, contact:
        edo.jelavic@mavt.ethz.ch}%
}

\newcommand{\vspacemargin}{-0.6cm}
\algnewcommand{\LeftComment}[1]{\Statex \(\triangleright\) #1}

\DeclareMathOperator{\atantwo}{atan2}

\begin{document}

\maketitle
\thispagestyle{empty}
\pagestyle{empty}

%%%%%%%%%%%%%%%%%%%%%%%%%%%%%%%%%%%%%%%%%%%%%%%%%%%%%%%%%%%%%%%%%%%%%%%%%%%%%%%%
\begin{abstract}
Planning for legged-wheeled machines is typically done using trajectory optimization because of many degrees of freedom, thus rendering legged-wheeled planners prone to falling prey to bad local minima. We present a combined sampling and optimization-based planning approach that can cope with challenging terrain. The sampling-based stage computes whole-body configurations and contact schedule, which speeds up the optimization convergence. The optimization-based stage ensures that all the system constraints, such as non-holonomic rolling constraints, are satisfied. The evaluations show the importance of good initial guesses for optimization. Furthermore, they suggest that terrain/collision (avoidance) constraints are more challenging than the robot model's constraints. Lastly, we extend the optimization to handle general terrain representations in the form of elevation maps. 
\end{abstract}
%
%%%%%%%%%%%%%%%%%%%%%%%%%%%%%%%%%%%%%%%%%%%%%%%%%%%%%%%%%%%%%%%%
\begin{acronym}
\acro{PRM}{Probabilistic Roadmap}
\acro{DoF}{Degrees of Freedom}
\acro{CDRM}{Contact Dynamic Roadmap}
\acro{SBP}{Sampling Based Planning}
\acro{MPC}{Model Predictive Control}
\acro{CoP}{Center of Pressure}
\acro{CoM}{Center of Mass}
\acro{TO}{Trajectory Optimization}
\acro{EE}{End-Effector}
\acro{NLP}{Nonlinear Program}
\acro{OCP}{Optimal Control Problem}
\acro{HP}{Hermite Parametrization}
\acro{MPC}{Model Predictive Control}
\acro{CAN}{Controller Area Network}
\acro{QP}{Quadratic program}
\acro{RT}{Real-time}
\acro{HEAP}{Hydraulic Excavator for an Autonomous Purpose}
\acro{LF}{Left Front}
\acro{RF}{Right Front}
\acro{LH}{Left Hind}
\acro{RH}{Right Hind}
\acro{WBC}{Whole Body Controller}
\acro{IK}{Inverse Kinematics}
\acro{AA}{Abduction/Adduction}
\acro{FE}{Flexion/Extension}
\acro{IMU}{Inertial Measurement Unit}
\acro{GNSS}{Global Navigation Satellite System}
\acro{RTK}{Real-Time Kinematic}
\acro{RBDL}{Rigid Body Dynamics Library}
\acro{SDF}{Signed Distance Function}
\acro{RRT}{Rapidly-Exploring Random Tree}
\acro{RS}{Reeds-Shepp}
\acro{NN}{Nearest Neighbor}
\end{acronym}
%%%%%%%%%%%%%%%%%%%%%%%%%%%%%%%%%%%%%%%%%%%%%%%%%%%%%%%%%%%%%%%%%%%
%

%%%%%%%%%%%%%%%%%%%%%%%%%%%%%%%%%%%%%%%%%%%%%%%%%%%%%%%%%%%%%%%%%%%%%%%%%%%%%%%%
%%%%%%%%%%%%%%%%%%%%%%%%%%%%%%%%%%%%%%%%%%%%%%%%%%%%%%%%%%%%%%%%%%%%%%%%%%%%%%%%
\section{INTRODUCTION}
\label{sec::introduction}
Equipping robots with legs enables mobility over unstructured terrains, the same as the animals can traverse them. On the other hand, wheeled robots are not as mobile as their legged counterparts, but they are far more energy-efficient. By combining legs and wheels into a \emph{hybrid system}, roboticists have tried to keep the best characteristics of legged and wheeled systems \cite{bjelonic2020rolling, sun2020towards, klamt2017anytime}. Unfortunately, increased flexibility comes with an increase in complexity, especially true for motion planning since the combinatorial aspect of legged robots (contact schedule) is combined with wheeled robots' non-holonomic nature (rolling constraints). A powerful tool used to cope with the increased complexity is \ac{TO}. \ac{TO} has been shown to work well on hybrid systems with many \acp{DoF} (\cite{bjelonic2020rolling,jelavic2019whole, du2020whole}). However, optimization is prone to local minima. While some control over solution quality can be achieved through cost function shaping (e.g., \cite{melon2020reliable,medeiros2020trajectory}), the optimization can ultimately fail due to the non-convexity of the planning problem. Failures happen especially often in challenging terrain where collision (avoidance) constraints become particularly difficult for optimizers. In this letter, we overcome this problem by providing good initializations (\emph{initialization step}) for the optimization-based planner (\emph{refinement step}); i.e we combine \ac{SBP} with \ac{TO}. Our \emph{initialization step} computes base poses, joint positions, and contact schedule, which are all then fed to the optimization. The gain is that our method can handle systems with many \ac{DoF} while still being able to cope with challenging terrain. 
\begin{figure}[tb]
\centering
    \includegraphics[width=1\columnwidth]{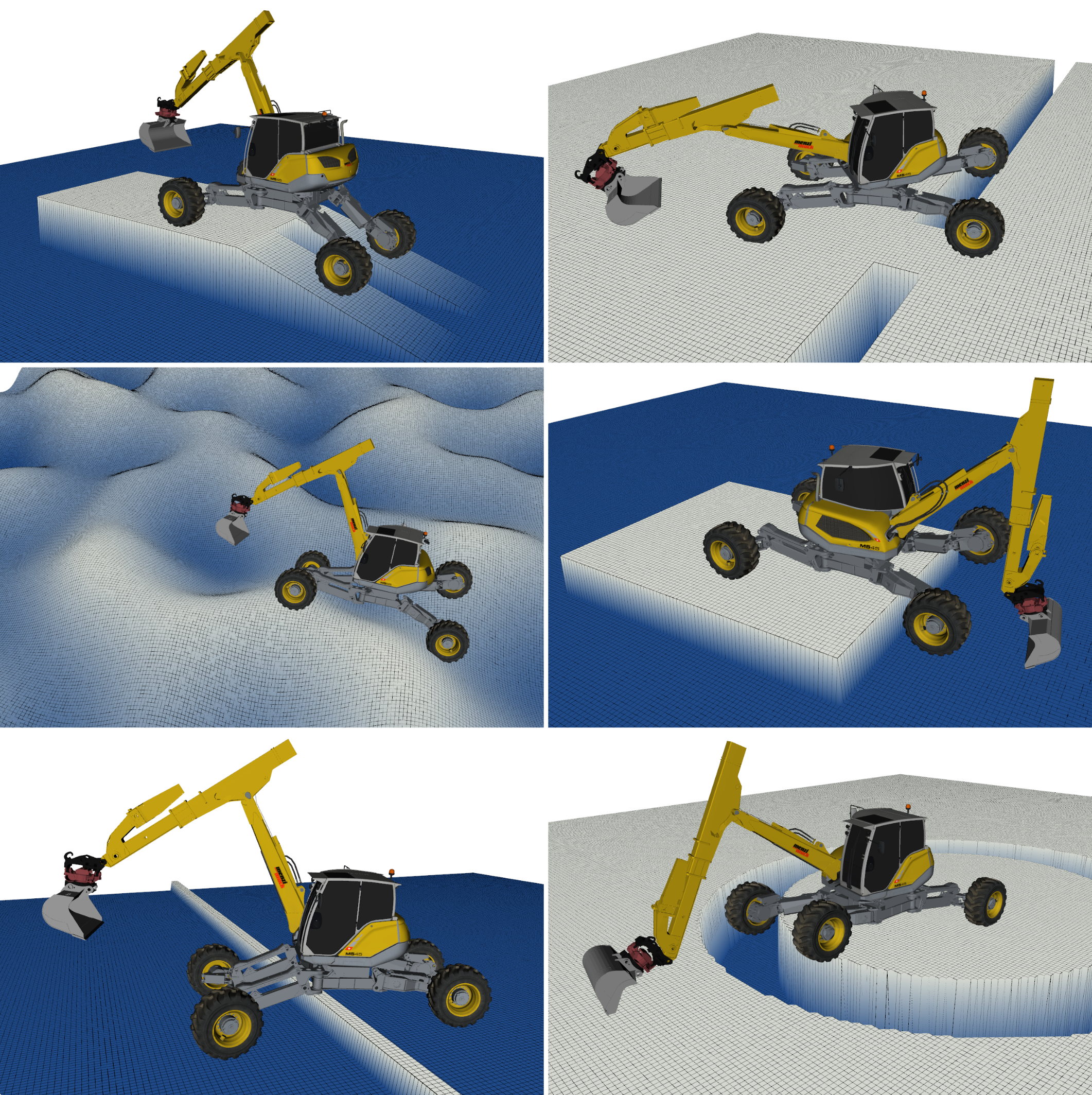}
    \vspace{-0.5cm}
    \caption{\ac{HEAP} navigating a variety of different terrains (refer to accompanying video for more examples). \textit{\textbf{Left-up:}} Driving up a \SI{2}{\meter} service ramp. \textit{\textbf{Right-up:}} \emph{hole} terrain, traversing over a hole via \SI{1}{\meter} wide passage which is too narrow for both legs. \textit{\textbf{Left-middle:}} \emph{rough} terrain, navigating a terrain with roughness of $\pm$\SI{1.5}{\meter}. \textit{\textbf{Right-middle:}} \emph{step} terrain, stepping on a \SI{1}{\meter} high block. \textit{\textbf{Left-down:}} Stepping over \SI{0.5}{\meter} high wall. \textit{\textbf{Right-down:}} \emph{gap} terrain, crossing a \SI{2}{\meter} wide gap.}
    \label{fig::terrains}
    \vspace{\vspacemargin}
\end{figure}
\subsection{Related Work}
Early research on locomotion for hybrid systems treats the whole system as a driving robot and uses legs as an active suspension. Such a practice is widespread in the aerospace community \cite{cordes2017static, giordano2009kinematic} and has more recently been applied to wheeled quadrupeds \cite{bjelonic2019keep}. The main disadvantage of such an approach is that it does not fully leverage the hybrid system's legged nature, i.e., the robot cannot negotiate obstacles.

To simplify the combinatorial nature of contact schedule planning for legged robots, researchers often use cyclic gaits to restrict the solution space size. Planning with cyclic gaits has been widely used for quadrupedal robots with point feet (\cite{bellicoso2018dynamic, winkler2017fast,rebula2007controller, farshidian2017real}) and more recently, it has been applied to hybrid robots as well (\cite{bjelonic2020rolling, de2019trajectory,du2020whole}). Presented controllers run under the flat ground assumption in an \ac{MPC} fashion with a prediction horizon of about \SI{2}{\second}. Such a strategy relies on reactive control behavior from \ac{MPC}, and while it can traverse small irregularities in the terrain, large obstacles still pose an issue. Furthermore, these approaches are not suitable for temporally global plans because of the relatively short prediction horizon.

More recently, terrain-aware planning has been proposed for hybrid robots (\cite{medeiros2020trajectory, sun2020towards}), which demonstrate the ability to traverse challenging terrain and plan motions in a whole-body fashion. However, the presented methods solely rely on trajectory optimization and often fall prey to the local minima. Furthermore, the environment is known \emph{excatly}, and it remains unclear how the planner would handle maps generated from real sensory data since discretization and noise can lead to discontinuous gradients in the optimization.

Unlike the optimization, \ac{SBP} cope well with the non-convex environment. Attempts to use \acp{SBP} can be found in \cite{tonneau2018efficient, geisert2019contact}, where the proposed approach samples base poses in $SE(3)$ space and computes a guiding path for the base of the robot. The footholds are computed in the next stage using the guiding path. However, the proposed approach does not use any optimization, which would make the planning for robots with non-holonomic constraints difficult.  Unlike \cite{tonneau2018efficient}, our approach does not accept a guiding path before having computed the footholds. 

Recently, \cite{short2017legged} has introduced a \ac{CDRM} data structure for rapid collision checking and foothold generation at runtime. The crux of the approach is computing a \ac{PRM} of collision-free configurations offline and then using it for planning online. A similar idea is employed for self-collision avoidance during the initialization phase in our work. We extend the \ac{PRM} with additional data, so that the approach is applicable to robots with heavy limbs such as walking excavators.

Finally, Klamt et al. \cite{klamt2017anytime}, \cite{klamt2018planning} use a graph-search based approach to plan hybrid motions for the Momaro robot. The robot decides when to drive and when to step based on a carefully crafted cost function. The transition sequences between driving and stepping are not computed in a whole-body fashion (as they are handcrafted). Lastly, the graph-search algorithm choice is motivated by the fact that the robot can turn in place using its wheels only. Hence, the approach is unsuitable for robots with a minimal turning radius greater than zero in its current form.
\subsection{Contribution}
We present a combined sampling and optimization based planner for legged-wheeled machines with many \acp{DoF}. We use terrain representation to generate a wide variety of locomotion behaviors for navigating complex terrain. The planner is divided into two stages: \emph{Initialization Step} based on a sampling-based planner and \emph{Refinement Step} through nonlinear optimization. These two stages produce kinematically feasible and statically stable plans, motivated by our use case on a walking excavator \cite{HEAP}. Given the base's initial and goal pose, our formulation computes base trajectory in $SE(3)$, joint trajectories, and contact schedule. To the best of the author's knowledge, existing approaches provide only high-level waypoints (e.g., \cite{wermelinger2016navigation,bellicoso2018advances,klamt2018planning}). Finally, our approach is the first (to the best of our knowledge) to include general terrain representations into an optimization-based planner for hybrid systems.
\section{PROBLEM STATEMENT}
\label{sec::problem_statement}
A legged-wheeled robot comprises $ N $ limbs with wheels and $M$ limbs without wheels, e.g., a walking excavator has four limbs with wheels, and one non wheeled limb (see \cite{HEAP}). The base of the robot can move in $SE(3)$ space. Robot's knowledge about the environment is contained in a map, which is a mapping $h: \mathbb{R}^2 \rightarrow \mathbb{R}$ that maps coordinates to various functions describing the environment (e.g., height, traversability). In this letter, a multilayered grid map \cite{Fankhauser2016GridMapLibrary} data structure is used.
Robot's $i^{th}$ limb is in contact with the environment if it is close enough to the surface, i.e.
\begin{align}
    |\boldsymbol{p}^z_i - h(\boldsymbol{p}^x_i,\boldsymbol{p}^y_i)| \leq \epsilon, \; \epsilon \in \mathbb{R}_+
\end{align}
where $\boldsymbol{p}^z$ denotes $z$ component of the contact position $\boldsymbol{p}$ and $h(\boldsymbol{p}^x,\boldsymbol{p}^y)$ is height at the contact position. We assume that the contact point is below the center of the wheel (negative gravity direction).
\begin{figure}[t]
\centering
\subfloat[Raw terrain \label{fig::small_block_no_fileter}]{
       \includegraphics[width=0.48\columnwidth]{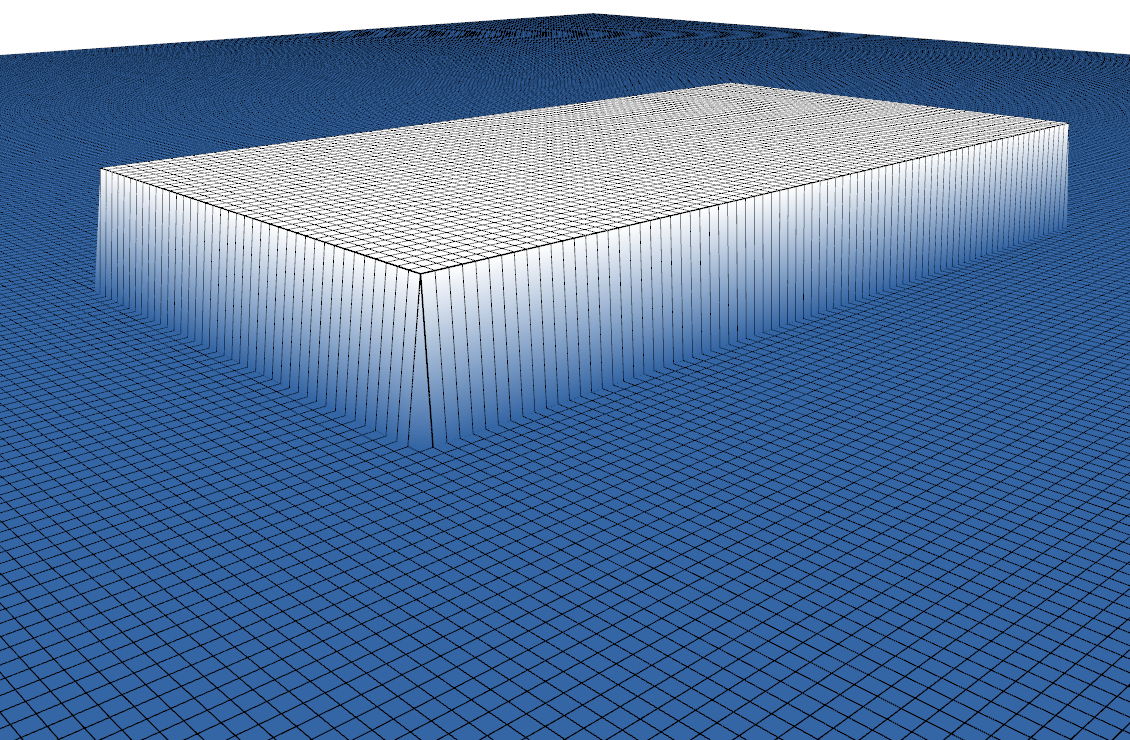}}
     \subfloat[Filtered terrain  \label{fig::small_block_filtered}]{
       \includegraphics[width=0.48\columnwidth]{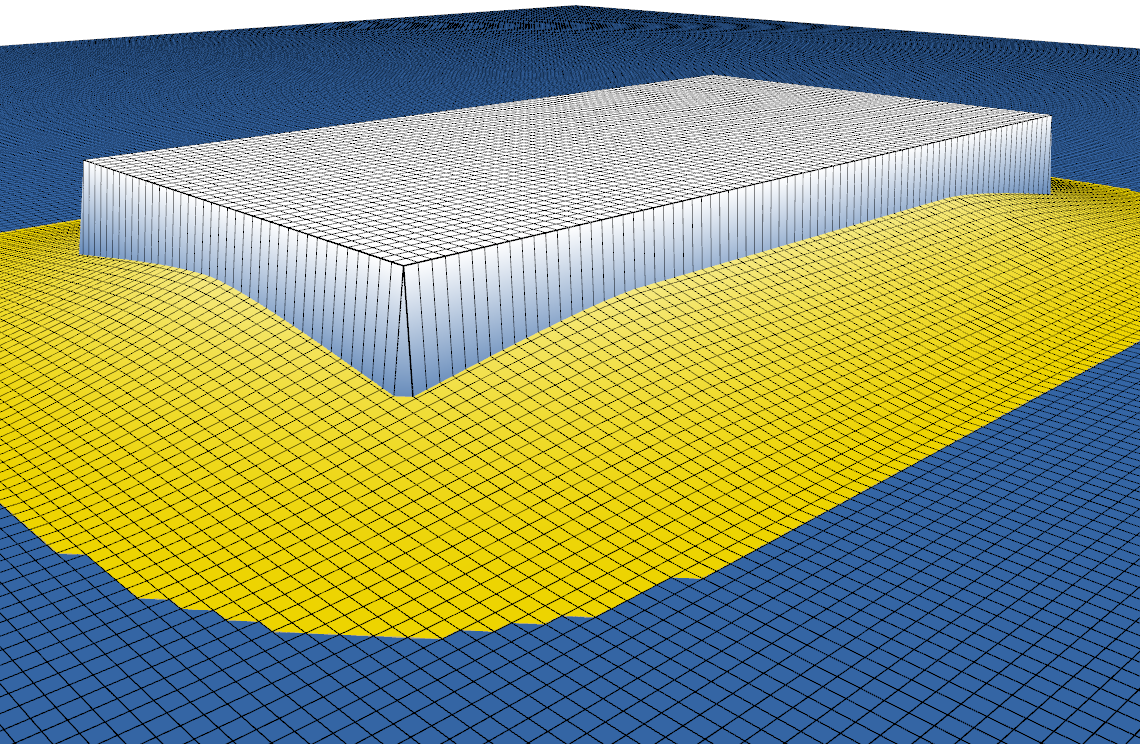}
     }
    \caption{\textbf{\textit{Left:}} Elevation map of a block with a height jump. The color gradually changes from blue to white, where blue areas are the lowest. \textbf{\textit{Right:}} Filtered elevation map shown in yellow color. Areas with dark yellow color have the lowest height. The filtered elevation can be used to compute base pose, as described in Sec.~\ref{sec::initialization_step}}
    \label{fig::small_block}
    \vspace{\vspacemargin}
\end{figure}
For a \SI{30}{\degree} slope (which is the limit beyond which we consider terrain untraversable), the approximation error is about \SI{15}{\percent} of the wheel radius, which is well within the adaptation capabilities of the tracking controller, as demonstrated in \cite{jelavic2020terrain}.

The environment is divided into \emph{traversable} part denoted with $\mathcal{T}$ and \emph{untraversable} part denoted with $\lnot\mathcal{T}$. A contact is valid if the contact point lies in the traversable set $\mathcal{T}$, i.e.
\begin{align}
\label{eq::traversability_constraint}
        (\boldsymbol{p}^x,\boldsymbol{p}^y) \in \mathcal{T} \equiv sdf(\boldsymbol{p}^x,\boldsymbol{p}^y)> \delta > 0
\end{align}
where $sdf(\cdot,\cdot)$ represents a 2D \ac{SDF} with the positive distance meaning the point lies in $\mathcal{T}$. We require all the contact points to stay at least $\delta$ away from the $\lnot \mathcal{T}$. The 2D \ac{SDF} is stored as a grid map layer and calculated using marching parabolas \cite{felzenszwalb2012distance}. Note that, unlike the limbs, robot's base is not required to stay in $\mathcal{T}$ since the base is not in contact with the terrain.

Our goal is to find a trajectory $\tau$ with a length of $T$ seconds such that $\boldsymbol{p}_{B}(t=0) = \boldsymbol{p}_{B,start}$ and $\boldsymbol{p}_{B}(t=T) = \boldsymbol{p}_{B,goal}$ where $\boldsymbol{p}_{B,start}, \boldsymbol{p}_{B,goal}$ are given starting and goal position for the base of the robot. Apart from the base, we do not enforce any other constraints on robot's pose or joint angles although this is not a hard requirement of our approach.
%
%
%
%
\begin{comment}
stuff not used (kicked out)

\begin{figure}[t]
\centering
\subfloat[Contact point \label{fig::contact_point}]{
       \includegraphics[width=0.32\columnwidth]{figures/wheel_contact_1_cropped.png}}
       \hspace{1cm}
     \subfloat[Approximation error  \label{fig::contact_point_schematic}]{
       \includegraphics[width=0.37\columnwidth]{figures/wheel_contact_2_cropped.png}
     }
    \caption{\textbf{\textit{Left:}} Contact point denoted with a red dot is assumed to be precisely under the center of the wheel (following the direction of gravity) \textbf{\textit{Right:}}  For the wheel of radius $R$ with the slope angle $\alpha$, the approximation error from our assumption amounts to $e$.}
    \label{fig::contact_points}
    \vspace{\vspacemargin}
\end{figure}

From the Fig.~\ref{fig::contact_point_schematic}, it can be seen that the following holds for the error $e$ in case of driving on a slope with inclination $\alpha$: 
\begin{align}
    cos(\alpha) = \frac{R}{R+e}
\end{align}
where $R$ is radius of the wheel and $e$ is the error of the approximation.
\end{comment}
%
%\input{preprocessing.tex}
%
%%%%%%%%%%%%%%%%%%%%%%%%%%%%%%%%%%%%%%%%%%%%%%%%%%%%%%%%%%%%%%%%%%%%%%%%%%%%%%%%
\section{INITIALIZATION STEP}
\label{sec::initialization_step}
The backbone of the initialization step is a sampling-based planner that samples base poses. We use \acp{RRT} \cite{karaman2011sampling, lavalle2006planning} although the problem formulation also permits the use of multi-query planners such as \acp{PRM} \cite{kavraki1996probabilistic}. Results presented are mostly generated using and RRT\# \cite{arslan2013use}, whose implementation is taken from \cite{sucan2012open}. In addition to \ac{RRT}, a \ac{PRM} of limb end-effector positions is precomputed for efficient online planning.  
\subsection{Offline Computation}
\begin{figure}[tb]
\centering
\subfloat[Roadmap legs \label{fig::roadmap_legs}]{
       \includegraphics[width=0.45\columnwidth]{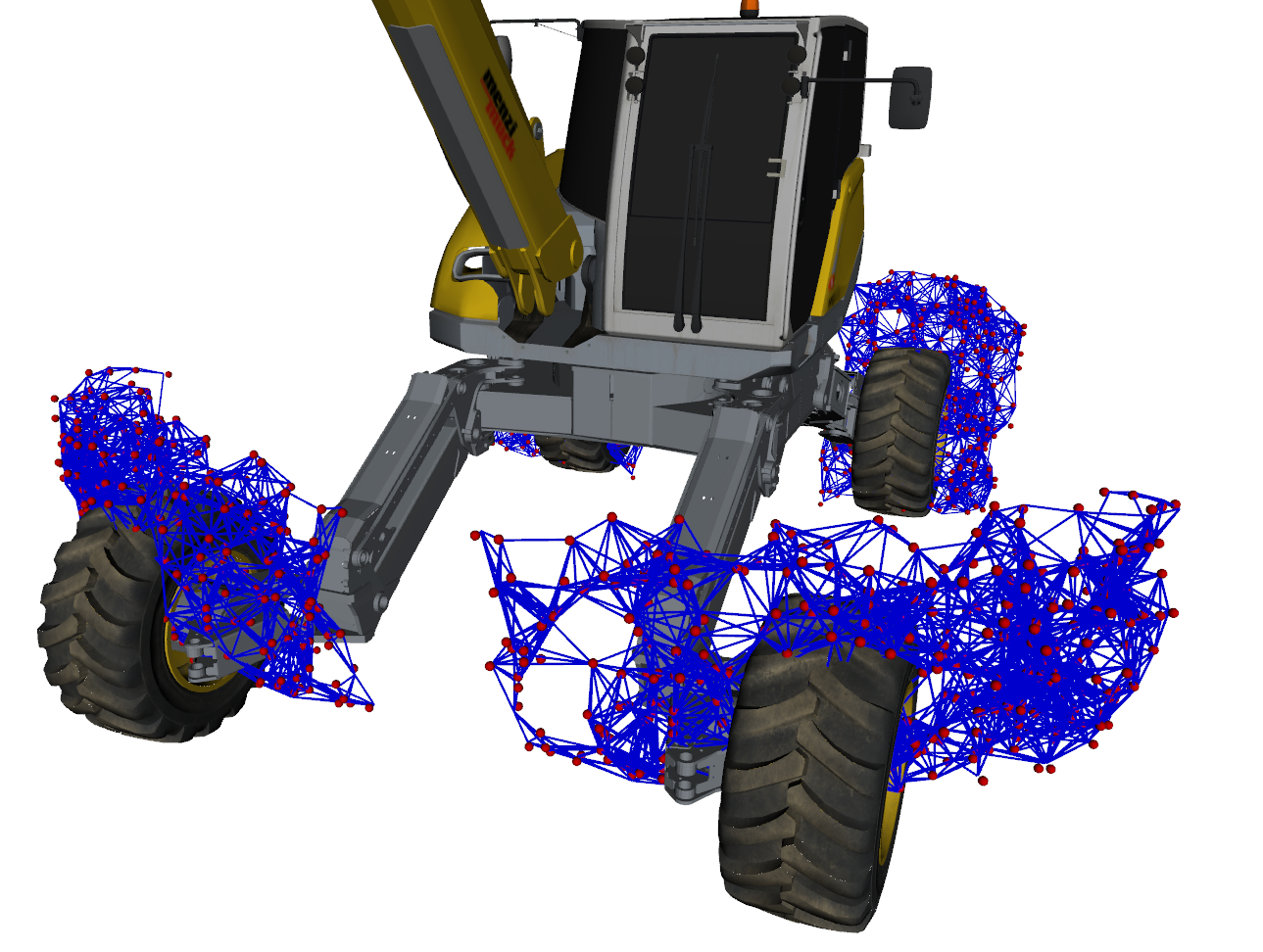}}
       \hfill
     \subfloat[Roadmap arm  \label{fig::roadmap_arm}]{
       \includegraphics[width=0.5\columnwidth]{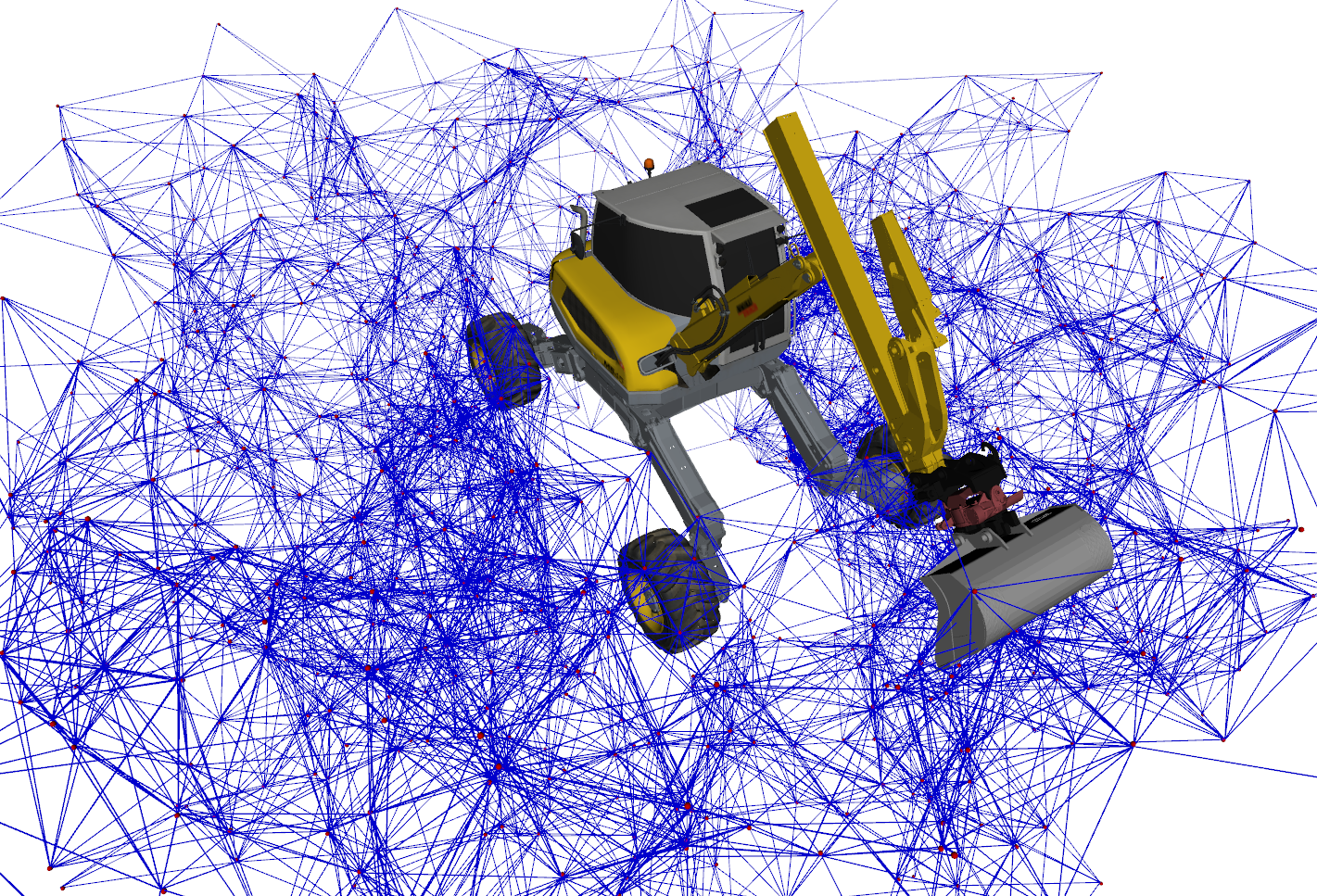}
     }
    \caption{Roadmap vertices (red spheres) and edges (blue lines). Each vertex represents an end-effector position in $\mathbb{R}^3$. \textbf{\textit{Left:}} Roadmap shown for the legs of a walking excavator. Shown are 300 vertices and about 2000 edges. \textbf{\textit{Right:}}  Roadmap for the arm. Shown are 1000 vertices and about 8000 edges.}
    \label{fig::roadmaps}
    \vspace{\vspacemargin}
\end{figure}

We use the \ac{CDRM} data structure introduced in \cite{short2017legged}. \ac{CDRM} can be used at runtime to generate collision-free movements of the robot's limbs. In our case, \ac{CDRM} helps us avoid collision between arm and legs while using the arm as a supporting limb. For each limb, the mapping between joint angles $\boldsymbol{q}_i$ and end-effector position $^{B} \boldsymbol{p_{i}}$ is stored. This mapping does not change over time since $^{B} \boldsymbol{p_{i}}$ is expressed in the base frame of the robot. Furthermore, we store the mapping between $i^{th}$ limb's \ac{CoM} $\boldsymbol{p}_{i,com}$ and joint configuration for the same limb. This allows us to evaluate the stability criterion during the online planning phase rapidly. The offline roadmap is created using the \ac{PRM} algorithm; it is shown in Fig.~\ref{fig::roadmaps}. The roadmap does not change over time which motivates the use of \ac{PRM}.
In addition to the roadmap, we compute terrain normals and filtered height. This computation can be performed by fitting a tangent plane at each point $(x,y)$ locally. Local fitting is done using least squares, and as a result, we get the normal of the plane and fitted height at point $(x,y)$. We filter once with a local radius $R$ of \SI{0.3}{\meter} and once with  \SI{2.5}{\meter}, which is roughly the robot's footprint radius. Filtering result  with large $R$ is shown in Fig.~\ref{fig::small_block}. The idea behind this is to use the filtered height for computing base poses. Terrain discontinuities (e.g., steps) should not be reflected in the base movement since the base moves above the terrain. On the other hand, smoothed terrain (shown in yellow) is a good approximation for base movement, assuming that it stays at some (roughly) constant height above the smoothed terrain.
%
%
%
% \begin{figure}[tb]
%     \centering
%     \includegraphics[width=1.0\columnwidth]{figures/wcrm_m545_2.png}
%     \caption{one variacnt}
%     \label{fig::cdrm}
% \end{figure}
%
%
% \begin{figure}[tb]
% \centering
% \subfloat[Roadmap legs \label{fig::roadmap_legs}]{
%       \includegraphics[width=0.45\columnwidth]{figures/wcrm_m545_2.png}}
%       \hfill
%      \subfloat[Roadmap arm  \label{fig::roadmap_arm}]{
%       \includegraphics[width=0.5\columnwidth]{figures/wcrm_arm_m545.png}
%      }
%     \caption{\textbf{\textit{Left:}} Roadmap shown for the legs of a walking excavator. Shown are 300 vertices (red spheres) and about 2000 edges (blue lines). \textbf{\textit{Right:}}  Roadmap for the arm. Shown are 1000 vertices and about 8000 edges.}
%     \label{fig::roadmaps}
% \end{figure}
%
%
%
\subsection{Online Planning}
Having computed the roadmap shown in Fig.~\ref{fig::roadmaps}, an \ac{RRT} planner finds a plan between base poses $\boldsymbol{p}_{B,start}$ and $\boldsymbol{p}_{B,goal}$. We use \ac{RRT} framework to enable re-planning in potentially changing maps. Similar to previous work \cite{tonneau2018efficient}, our planner proposes a base pose before computing the limb contacts. Subsequently, we check whether contacts can be established and whether the robot is stable. In general, a sampling-based planner typically has three main components: sampling, connecting a new sample to the tree, and feasibility checking. In this section, we describe how each step works.

\subsubsection{Sampling} \label{sec::sampling}Candidate base poses are sampled in $SE(2)$ space instead of full-fledged $SE(3)$. The idea behind this decision is straightforward: since limbs interact with the environment, the sampler uses terrain information to constrain some \acp{DoF} of the base pose, which reduces the dimension of the search space. Hence our planner samples $x,y$ position of the base, and yaw angle $\gamma$ from a uniform distribution. The remaining \acp{DoF} are computed based on local terrain features: roll angle $\alpha$, pitch angle $\beta$, and $z$ coordinate. Roll and pitch are computed from terrain normal $\boldsymbol{n}$ such that the base remains roughly parallel to the terrain underneath.
%
\begin{comment}
Roll and pitch at coordinate $\boldsymbol{p} = (x,y)$ can then be computed from the normal vector $\boldsymbol{n}$:
\begin{equation}
    \label{eq::roll_pitch_simple}
    \alpha = \atantwo (\boldsymbol{n}^y,\boldsymbol{n}^z), \;\; \beta= \atantwo (\boldsymbol{n}^x,\boldsymbol{n}^z)
\end{equation}
\end{comment}
%
Finally the $z$ coordinate can be computed as $z = h(x,y) + h_{desired}$ where $h(x,y)$ is the terrain elevation at sampled point $(x,y)$ and $h_{desired}$ is user defined desired height above the terrain. 

Selecting  $h(x,y)$ and $\boldsymbol{n}$ can be tricky. E.g., when crossing a deep gap, terrain height can be so low that the planner cannot generate any valid poses (despite the heavy filtering). Luckily, one can leverage a simple observation to chose good $h$ and $\boldsymbol{n}$. When moving over untraversable terrain $\lnot \mathcal{T}$, the robot only cares about the nearest $\mathcal{T}$ (traversable area). The rationale is that contacts should only be made with $\mathcal{T}$, and the base pose should be selected such that limbs can reach the nearest $\mathcal{T}$. Alg.~\ref{alg::pose_generation} implements this proposition; it selects $h$ and $\boldsymbol{n}$ such that contacts with $\mathcal{T}$ can be established.
\begin{algorithm}[t]
 \small 
  \caption{Select $\boldsymbol{n}$ and $h$ at position $(x,y)$}
  \label{alg::pose_generation}
  \begin{algorithmic}[1]
    \LeftComment {Input: base position $(x,y)$, grid map}
    \State $(\hat{x},\hat{y})$ = nearestTraversablePosition$(x,y)$
    \State $h$ = height$(x,y)$,  $\boldsymbol{n}$ = normal$(x,y)$
    \State $h_{f}$ = heightFiltered$(x,y)$, $\boldsymbol{n}_{f}$ = normalFiltered$(x,y)$
    \State $\hat{h}$ = height$(\hat{x},\hat{y})$, $\hat{\boldsymbol{n}}$ = normal$(\hat{x},\hat{y})$
    \State $\hat{h}_{f}$ = heightFiltered$(\hat{x},\hat{y})$, $\hat{\boldsymbol{n}}_{f}$ = normalFiltered$(\hat{x},\hat{y})$
    \State Hs = \{ $h$, $h_{f}$, $\hat{h}$, $\hat{h}_{f}$\}, Ns = \{ $\boldsymbol{n}$, $\boldsymbol{n}_{f}$, $\hat{\boldsymbol{n}}$, $\hat{\boldsymbol{n}}_{f}$\}
    \For{$(\boldsymbol{n}_c, h_c)$ in $\{ Ns \times Hs \}$}
        \State $cost$ = computePoseCost($h_c$,$\boldsymbol{n}_c$)
            \If {$cost < c_{best}$}
                \State $h_{best}$ = $h_c$, $\boldsymbol{n}_{best}$ = $\boldsymbol{n}_c$, $c_{best}$ = $cost$
            \EndIf
    \EndFor 
    \State \Return ($h_{best}$, $\boldsymbol{n}_{best}$)
  \end{algorithmic}
\end{algorithm}
The Alg.~\ref{alg::pose_generation} enumerates all candidate normals and heights (lines 1-6) and then does a small brute force search to select a pair $(h_{best}, \boldsymbol{n}_{best})$ that minimizes some criterion. The \emph{computePoseCost} function in line 9, gives low cost to poses where all legs are grounded and penalizes big roll and pitch angles. Full base pose is then determined from $(h_{best}, \boldsymbol{n}_{best})$.
\subsubsection{Connection to Tree}Upon drawing a random base pose $(x,y,\gamma)$, the planner tries to connect it to the tree (using the weighted cost of euclidean distance and angular distance). The connection is done in $SE(2)$ space using \ac{RS} curves \cite{reeds1990optimal}. \ac{RS} curves give an optimal path between two poses while respecting the minimum turning radius constraint. For a robot that can turn in place, one could use a very small turning radius. Attempting straight-line connections between base poses would make the subsequent refinement step (see Sec.~\ref{sec::refinement}) very hard since the satisfaction of the non-holonomic rolling constraint cannot be guaranteed. By using \ac{RS} curves, this is implicitly ensured; however, the resulting trajectory might be longer. Computing the \ac{RS} connection is done terrain agnostic completely.
\begin{figure}
    \centering
    \includegraphics{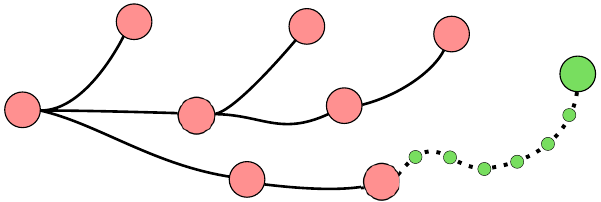}
    \caption{Addition of a new node into the \ac{RRT} tree. Red circles and full black lines are nodes and paths that make the current RRT tree. The newly sampled node (green) has to be connected to the rest of the tree. For a successful connection, all subnodes on the connecting path (small green circles) have to be valid.}
    \label{fig::rrt_step}
    \vspace{-0.2cm}
\end{figure}
\subsubsection{Feasibility Checking}
Fig.~\ref{fig::rrt_step} depicts the feasibility checking. Upon drawing a new sample (large green node) as described in step 1, we compute a \ac{RS} connection (dotted line) to the new node, as described in step 2. Subsequently, the dotted line is discretized into subnodes (small green circles) using an \ac{RS} interpolation method \cite{sucan2012open}. Next, for each subnode we generate full 6 \ac{DoF} pose using Alg.~\ref{alg::pose_generation}. Finally, each subnode undergoes feasibility checking, ensuring that the robot is statically stable and can establish enough contacts with the ground. In case a feasibility check passes for every subnode, the \ac{RRT} adds the new state and the connecting path to the tree. It is crucial to discretize the \ac{RS} path with high resolution since straight-line connections are assumed between two subsequent subnodes; in our implementation, we allow for a maximal distance of \SI{20}{\centi \meter}. The length of the whole path is a tuning parameter (in our case, \SI{15}{\meter}). Alg.~\ref{alg::feasibility_check} summarizes feasibility checking.
%
%
% New definitions
\algnewcommand\algorithmicswitch{\textbf{switch}}
\algnewcommand\algorithmiccase{\textbf{case}}
\algnewcommand\algorithmicassert{\texttt{assert}}
\algnewcommand\Assert[1]{\State \algorithmicassert(#1)}%
% New "environments"
\algdef{SE}[SWITCH]{Switch}{EndSwitch}[1]{\algorithmicswitch\ #1\ \algorithmicdo}{\algorithmicend\ \algorithmicswitch}%
\algdef{SE}[CASE]{Case}{EndCase}[1]{\algorithmiccase\ #1}{\algorithmicend\ \algorithmiccase}%
\algtext*{EndSwitch}%
\algtext*{EndCase}%
\begin{algorithm}[t]
    \small 
  \caption{Check feasibility of base pose $\boldsymbol{T}$}
  \label{alg::feasibility_check}
  \begin{algorithmic}[1]
    \State $(\alpha,\beta)$ = rollAndPitch$(\boldsymbol{T})$
    \If{  $|\alpha| > \alpha_{max}$ or $|\beta|>\beta_{max}$ }
        \State \Return FALSE
    \EndIf
    \State $nContacts$ = selectContactLegsConfiguration($\boldsymbol{T}$)
    \Switch{$nContacts$}
        \Case{$4$}
            \State \Return TRUE
        \EndCase
        \Case{$3$}
            \State selectSwingLegsCofiguration($\boldsymbol{T}$)
            \State selectSwingArmCofiguration($\boldsymbol{T}$)
            \State \Return isStable()
        \EndCase
        \Case{$2$}
            \State selectContactArmConfiguration($\boldsymbol{T}$)
            \State selectSwingLegsCofiguration($\boldsymbol{T}$)
            \State \Return isStable()
        \EndCase
        \Case{$1,0$}
            \State \Return FALSE
        \EndCase
    \EndSwitch
  \end{algorithmic}
\end{algorithm}

The feasibility check shown in Alg.~\ref{alg::feasibility_check} does not allow poses with a large roll or pitch angle (line 2). We then ground all the legs (line 5). A leg is grounded if a sufficient number of configurations in \ac{PRM} are in contact with the surface and the contact location lies in $\mathcal{T}$ (see Sec.~\ref{sec::problem_statement}). Among grounded legs configurations the algorithm picks the one closest to the default configuration. In case of four contact legs, the pose is deemed to be stable. In case three legs are in contact, the algorithm selects good joint configuration for the swing leg according to some criterion (e.g. ground clearance or proximity to the default configuration). The swing arm configuration is selected such that the \ac{CoM} is as centralized as possible. In case only two legs are in contact, the algorithm checks whether the arm can be grounded (line 14) and then proceeds with selecting swing leg configurations. The \emph{isStable()} function computes the \ac{CoM} of the whole robot and verifies that it lies in the support polygon. Aside from performing feasibility checking, Alg.~\ref{alg::feasibility_check} computes the full joint state of the robot $\boldsymbol{q}$ for feasible poses. The \ac{CoM} of the full joint configuration $\boldsymbol{q}$ can be computed using:
\begin{equation}
    \label{eq::com}
    \boldsymbol{p}_{com}(\boldsymbol{q}) = \frac{1}{M} \Bigg( m_B\boldsymbol{p}_{B,com} + \sum_{i=1}^{N} m_i\boldsymbol{p}_{i,com}(\boldsymbol{q}_i) \Bigg)
\end{equation}
where $M$ is the mass of the whole robot, $\boldsymbol{p}_{i,com}$ and $m_i$ are the position of \ac{CoM} and mass of the $i^{th}$ limb, respectively. Thanks to the mapping between $\boldsymbol{q}_i$ and $\boldsymbol{p}_{i,com}$ computed offline, the sum in Eq.~\ref{eq::com} can be evaluated rapidly. Note that unlike \cite{tonneau2018efficient}, we do not require all limbs to contact the environment while generating base poses, thus allowing for more flexibility. 
\begin{figure*}[t]
\centering
     \subfloat[Costs, Time until first solution  \label{fig::evaluations_cost}]{
       \includegraphics[width=0.32\textwidth]{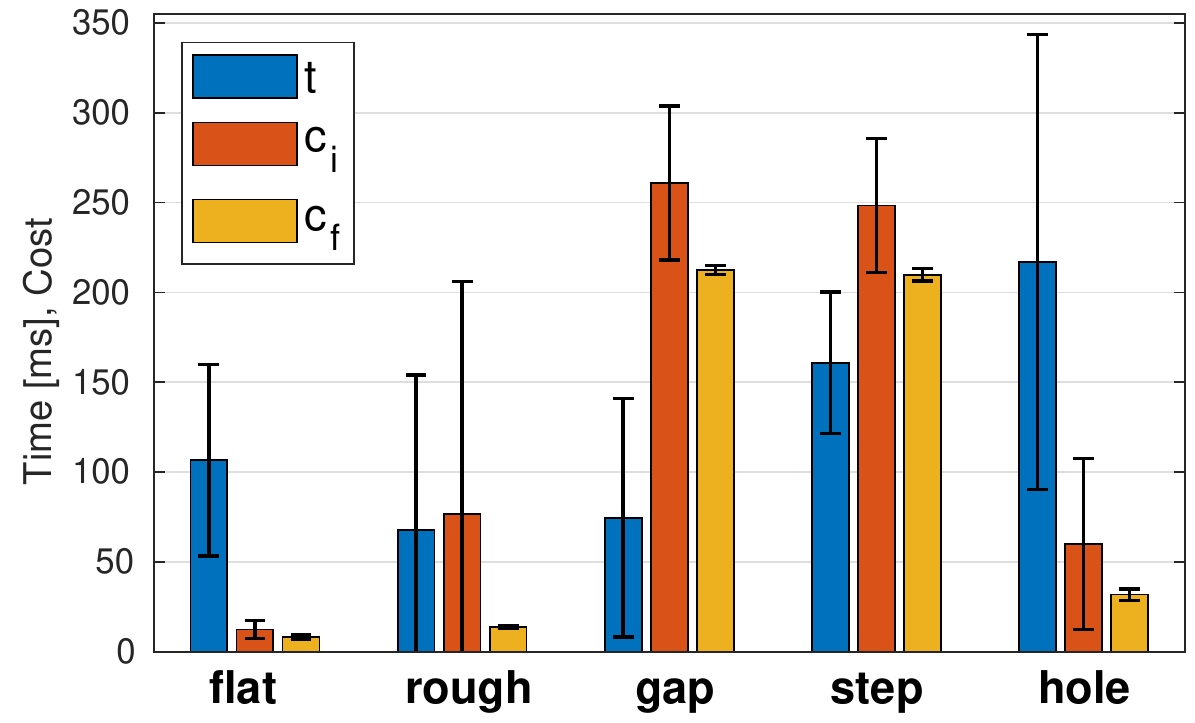}
     }
     \subfloat[Success rates \label{fig::evaluations_success}]{
       \includegraphics[width=0.32\textwidth]{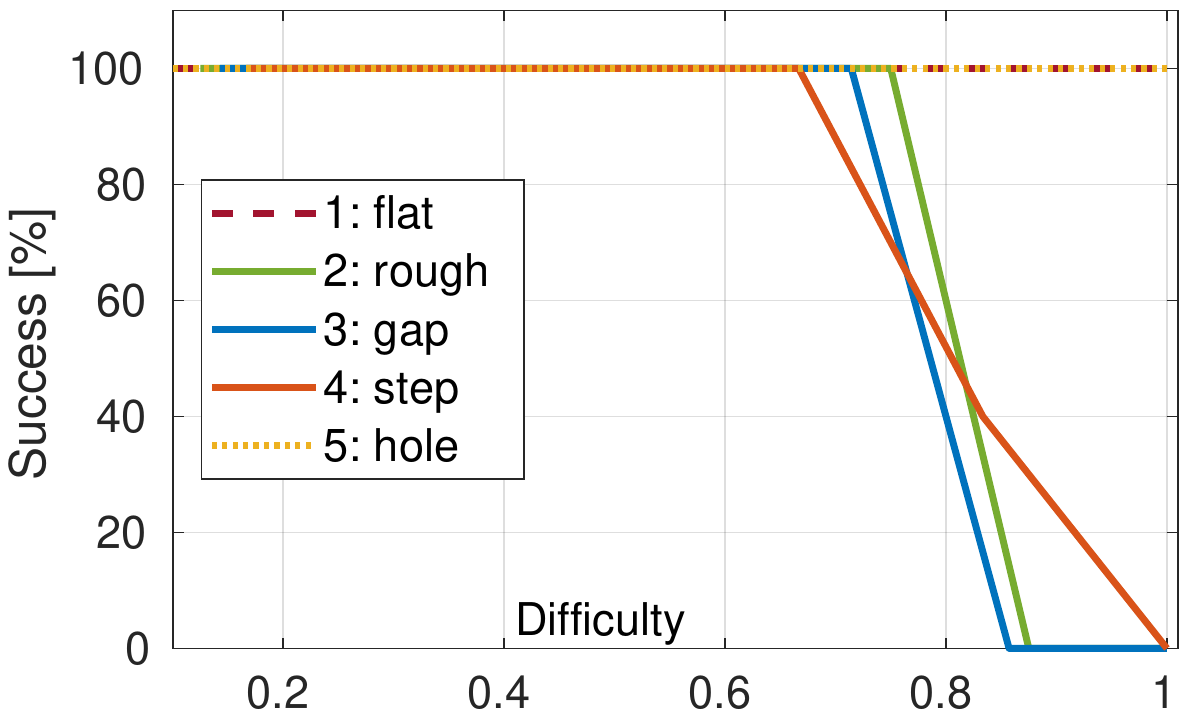}
     }
     \subfloat[Optimization times \label{fig::evaluations_optimization}]{
       \includegraphics[width=0.32\textwidth]{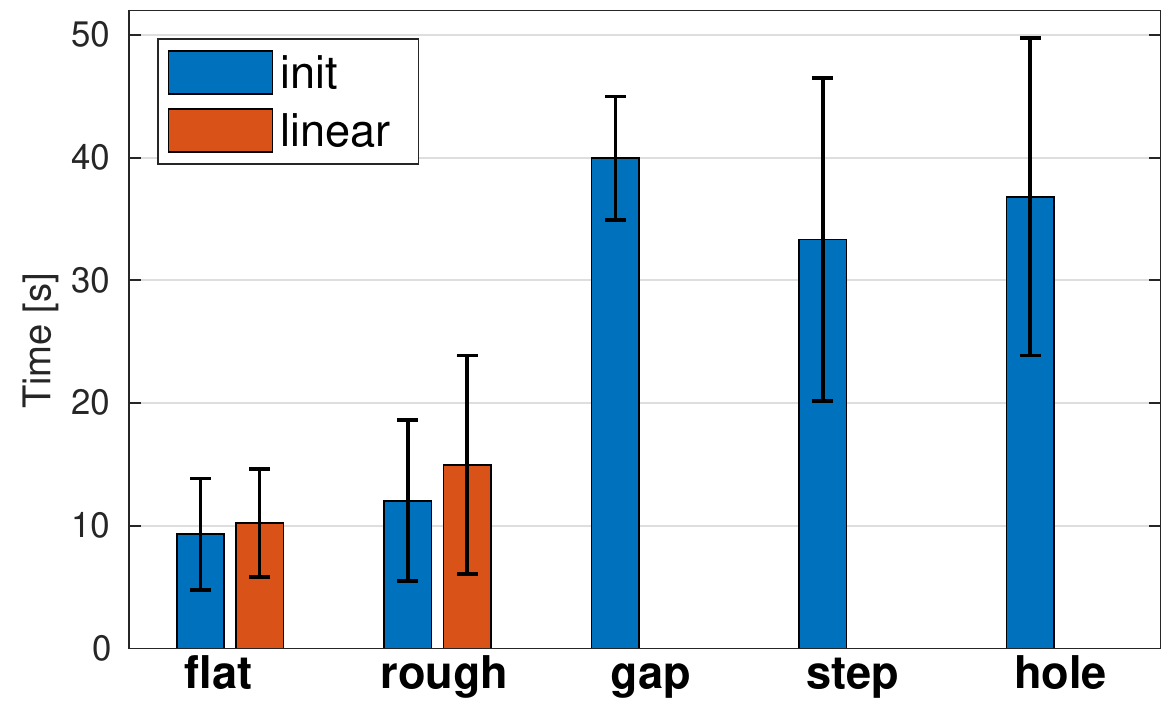}}
    \caption{Evaluation metrics \textbf{\textit{Left:}} Costs and computation times of the \ac{RRT} (optimization excluded). $t$-time until first solution  (blue), $c_i$-initial solution cost (red), $c_f$-final solution cost (yellow). \textbf{\textit{Middle:}} Success rates for different terrains versus difficulty. Maximal difficulty of 1, corresponds to roughness of $\pm2$ m, gap width of \SI{3.5}{\meter} and step height of \SI{1.5}{\meter}. Minimal difficulty corresponds to roughness of $\pm0.25$ m, gap width of \SI{1}{\meter} and step height of \SI{0.5}{\meter}. \ac{HEAP} wheel radius is \SI{0.6}{\meter} for comaprison. For the \emph{hole} terrain and flat terrain, the difficulty remains unchanged and the curves are shown for the sake of completeness. \textbf{\textit{Right}} Computation time until optimization convergence when initializing using planner's first stage versus using linear interpolation. For terrains with obstacles, \emph{gap}, \emph{step} and \emph{hole}, linear interpolation cannot find a solution.
    }
    \label{fig::evaluations}
    \vspace{\vspacemargin}
\end{figure*}
Once the \ac{RRT} has reached $\boldsymbol{p}_{B,goal}$, the final path is post-processed. In the first step, the contact schedule is modified to ensure stability. We do not allow establishing/breaking more than one contact between two different successive nodes. If the robot wants to change more than one contact state at any point, we insert a node in between. Those situations happen only when the arm contact is established/broken. The robot tries to change the contact state of the arm and leg(s) simultaneously. We add a short full contact phase (legs + the arm) in between to ensure static stability. 
Secondly, we compute \ac{IK} for the non-wheeled limbs in contact. Any of those limbs has to satisfy the contact constraint $\dot{\boldsymbol{p}}_i = \boldsymbol{0}$. For each non-wheeled limb in contact, we find base poses at the beginning and the end of its respective contact phase. The reference position $\boldsymbol{p}^*$ for the \ac{IK} is found by solving:
\begin{equation}
      \min ||\boldsymbol{p}_s - \boldsymbol{p}_e||
\end{equation}
and setting $\boldsymbol{p}^* = (\boldsymbol{p}_{s}^*+\boldsymbol{p}_{e}^*)/2.0$, where $\boldsymbol{p}_s$ are all positions in the $i^{th}$ limb roadmap at the beginning of the contact phase and $\boldsymbol{p}_e$ at the end of the contact phase. We then compute $i^{th}$ limb's joint angles for every contact node as $\boldsymbol{q}_i = IK(\boldsymbol{p}^*)$.

Alg.~\ref{alg::feasibility_check} ensures that the robot is stable and that limbs are not in a collision. Nevertheless, it assumes that straight line connections in joint space are collision-free. The assumption might be invalid, especially for the arm, which moves around the base and is used as a counterweight. To overcome this problem, we use the precomputed roadmap in which we, similar to \cite{short2017legged}, invalidate all vertices and edges that are in a collision with other limbs or the environment. Once the graph is updated, each limb's path is found using a graph search algorithm (A* in our case). In practice, legs always end up being collision-free, but the arm often collides with legs. Hence we run the graph search only between nodes where the arm moves.
%
%
\begin{comment}
State which of the shelf planner you use (show maybe some results with RRTstar, RRTsharp, BITstar). Say that you could use PRM as well. 

Need to describe the post processing

Need to describe the interpolation.

Need to describe sampling.

Put an image where you show the interpolation scheme. 
\end{comment}
%
%
%%%%%%%%%%%%%%%%%%%%%%%%%%%%%%%%%%%%%%%%%%%%%%%%%%%%%%%%%%%%%%%%%%%%%%%%%%%%%%%%
\section{REFINEMENT STEP}
\label{sec::refinement}
The refinement step uses \ac{TO} to satisfy all system constraints. \ac{TO} methods scale well with system dimension and can handle nonlinear constraints such as forward kinematics or non-holonomic rolling constraints. However, computing the correct contact schedule and dealing with obstacles remains challenging for the gradient-based methods, so we initialize optimization with trajectory computed in the \emph{initialization step}. The optimization receives contact schedule, base position/velocity (6 \ac{DoF}) and joint position/velocity (25 \ac{DoF} in our case) and solves a feasibility problem. Adding an optimization objective allows for fine motion tuning, however it typically results in increased computation times. \ac{TO} planner used in this paper is based upon our previous work \cite{jelavic2019whole}, and below we present modifications that enable us to cope with more challenging scenarios.

\emph{Terrain maps} To the best of the author's knowledge, there is no optimization-based planner for hybrid systems that can handle general terrain representations. Compared to their legged counterparts, hybrid robots keep their limbs in contact over long distances. Hence, map errors influence more variables and constraints, which makes the optimization more sensitive. So far, proposed terrain-aware optimization planners have used analytical descriptions of the environment \cite{medeiros2020trajectory}, \cite{winkler2018gait}, \cite{sun2020towards}. Our planner integrates grid maps \cite{Fankhauser2016GridMapLibrary} into the optimization, thus allowing planning in any environment where a 2.5D map is a suitable representation. The elevation map comes into the optimization in the form of height constraint for all limbs in contact $\mathcal{C}$.
\begin{equation}
\label{eq::height_constraint}
    \boldsymbol{p}_i^z = h(\boldsymbol{p}_i^x,\boldsymbol{p}_i^y), \; \forall i \in \mathcal{C}
\end{equation}
\begin{figure*}[t]
\centering
\subfloat[Prefer Shortest Path \label{fig::contact_happy}]{
       \includegraphics[width=0.24\textwidth]{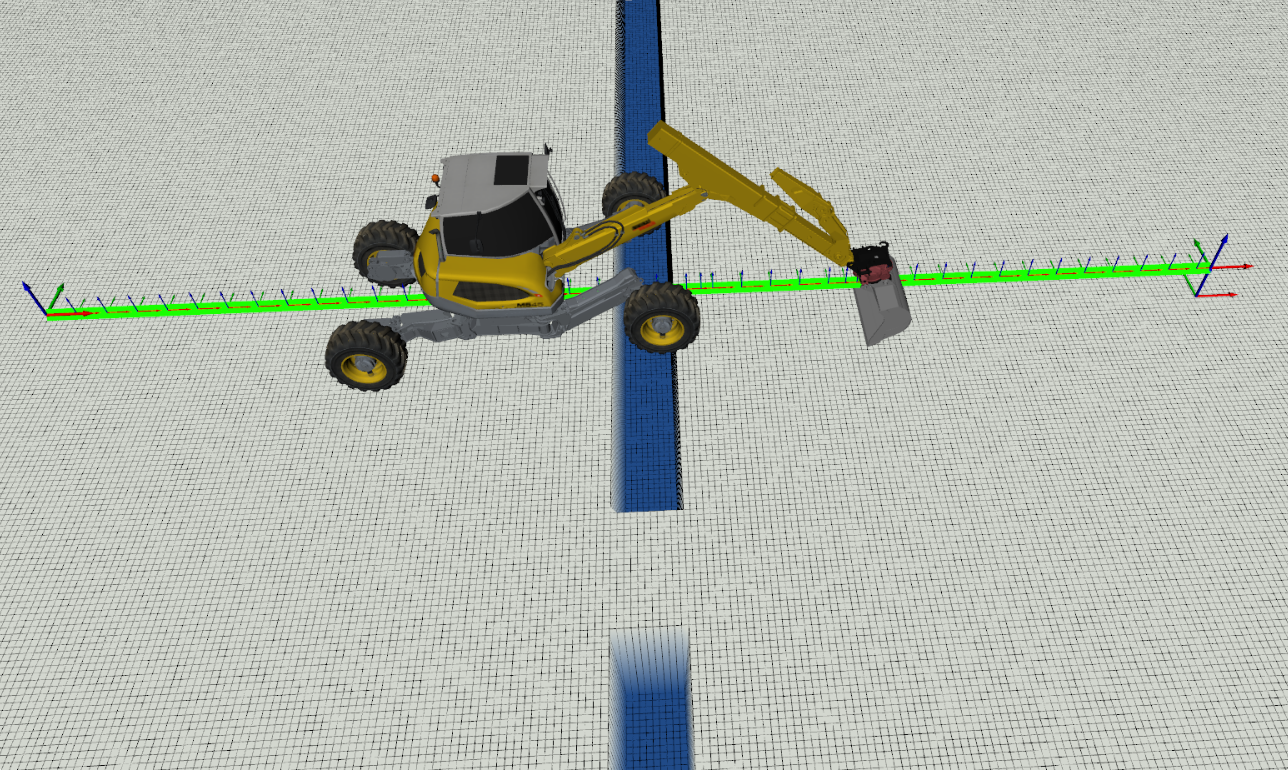}}
     \subfloat[Contact Schedule Stepping  \label{fig::contact_happy_schedule}]{
       \includegraphics[width=0.24\textwidth]{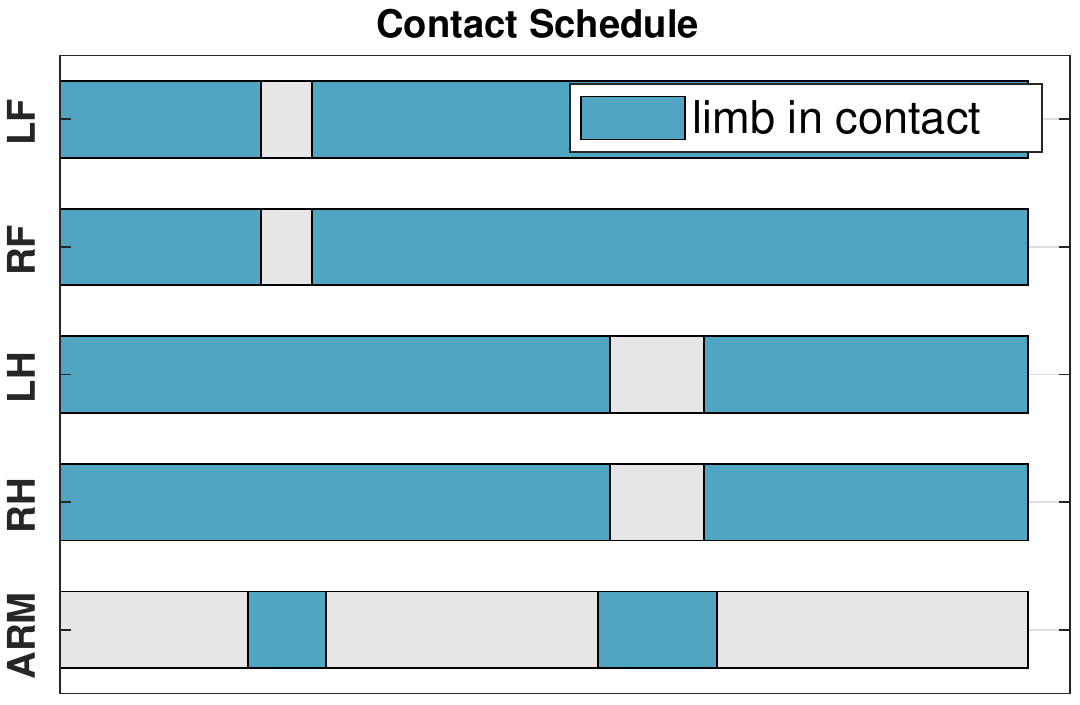}
     }
     \subfloat[Prefer Driving  \label{fig::contact_averse}]{
       \includegraphics[width=0.24\textwidth]{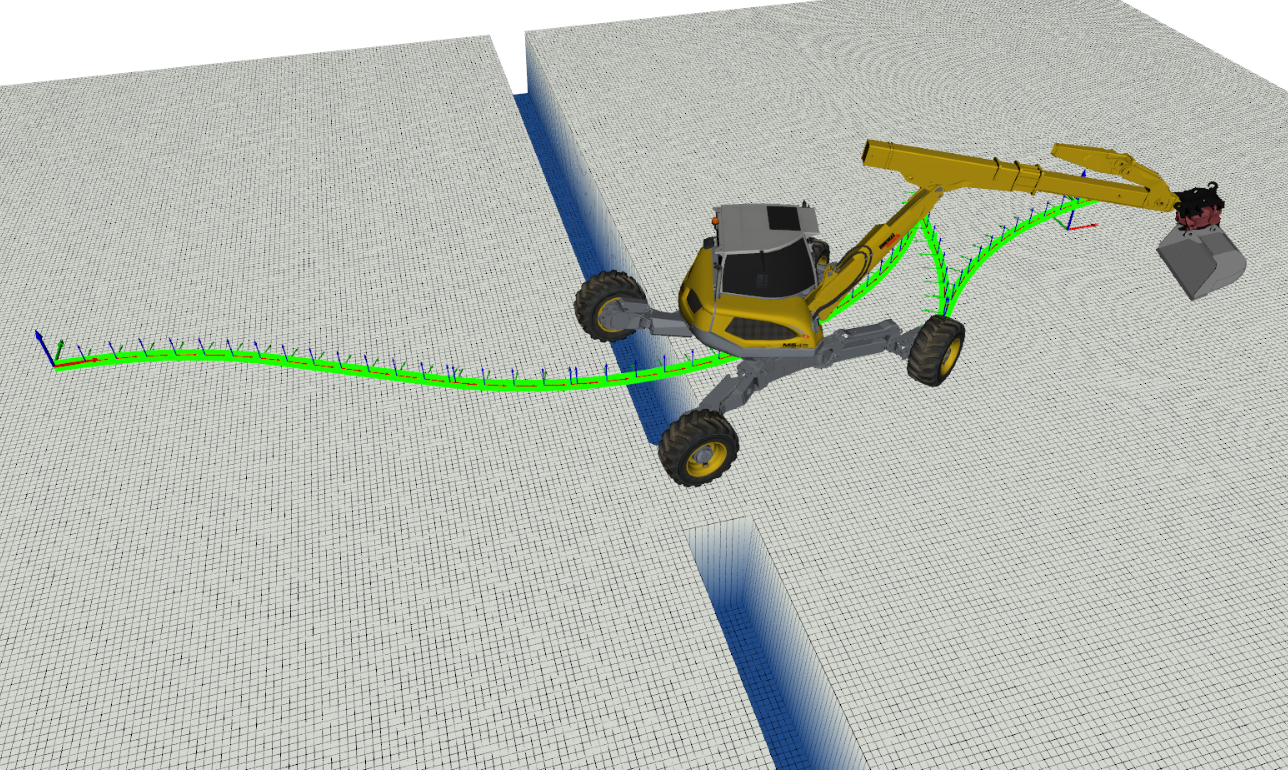}
     }
     \subfloat[Contact Schedule Driving  \label{fig::contact_averse_schedule}]{
       \includegraphics[width=0.24\textwidth]{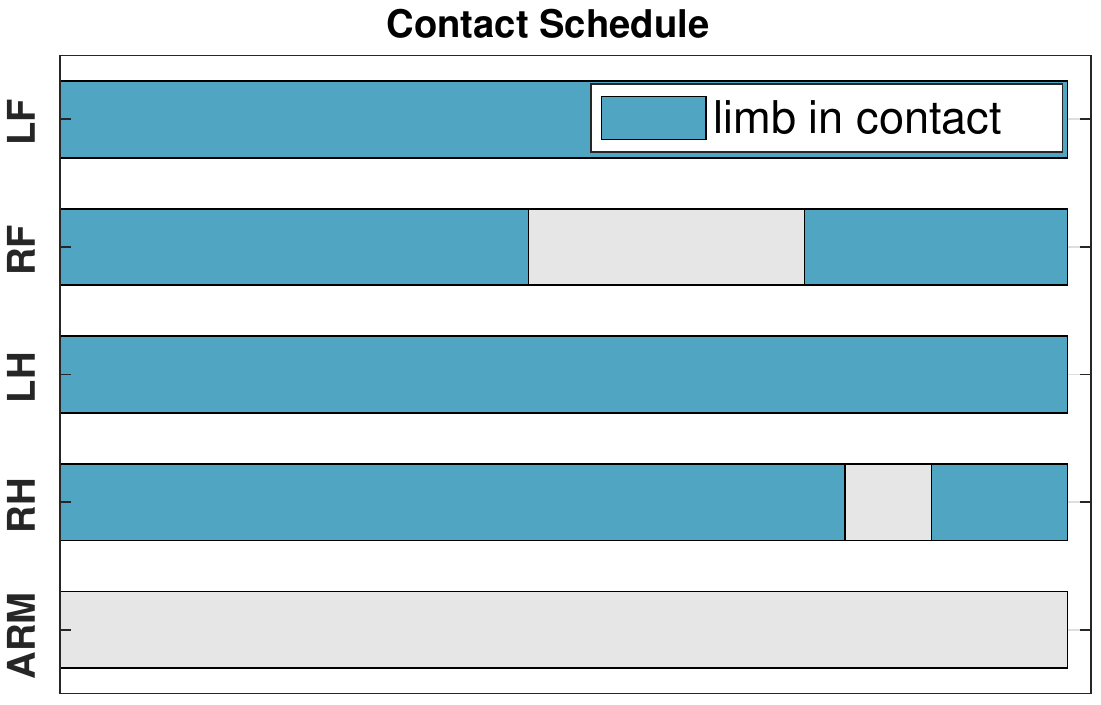}
     }
    \caption{Traversing terrain in different manners. Limb names: \textit{LF} stands for Left Front, whereas \textit{RH} stands for Right Hind etc. \textbf{\textit{Left:}} \ac{HEAP} reaches goal pose via shortest path \textbf{\textit{Middle Left:}} Contact schedule traversing over the gap. Total duration: \SI{114.98}{\second} \textbf{\textit{Middle Right:}} \ac{HEAP} trying to minimize stepping while reaching the goal. Not that two legs still have to be lifted since the bridge is too narrow for both legs. \textbf{\textit{Right:}} Contact schedule for driving whenever possible. Total duration: \SI{156.33}{\second}}
    \label{fig::locomotion_shaping}
    \vspace{\vspacemargin}
\end{figure*}
The constraint \eqref{eq::height_constraint} is problematic since height mapping $h(\cdot,\cdot)$ is discrete and discontinuous (e.g. gaps or steps) which can cause optimization to diverge. Far away from the cell center, \ac{NN} search or linear interpolation are poor approximations of the true elevation. Hence, approximation of partial derivatives $[\frac{\partial h}{\partial x} \; \frac{\partial h}{\partial y}]$ using central finite differences renders them non smooth. For large grid cells (\SI{0.1}{\meter}), the key to improving the optimization convergence is using a higher-order approximation of the $h(\cdot,\cdot)$ function. We found that bicubic interpolation and bicubic convolution algorithms \cite{keys1981cubic} work well. Implementations of both algorithms are integrated into the open-source package \emph{Grid Map} and made available for the community\footnote{\url{https://github.com/ANYbotics/grid_map}}. Unfortunately, neither filtering nor higher-order approximation can help if the terrain is discontinuous (steps or gaps). To handle discontinuities, we use \emph{gradient clipping}, a technique known from the machine learning community. Smaller clipping thresholds prevent getting stuck in bad minima. However, they usually require a few more iterations for convergence.

\emph{Traversability Constraint} We require that all contacts stay in the traversable area; the constraint implements Eq.~\ref{eq::traversability_constraint}. The gradient is also computed using a central finite difference with bicubic interpolation.

\emph{Collision Avoidance Constraint} imposes a minimum distance between the collision geometries of the robot. We use shape primitives such as spheres or cylinders. We impose that the minimal distance between collision geometries has to be greater than $d_{min}$. The minimal distance between collision geometries is calculated using the \emph{Bullet} physics engine. 
%
%
\begin{comment}
Here introduce the previous work and say how you extend the planner to incorporate collision (avoidance) constraints and traversability constraints.
It would be good to quantify timing how initialization affects timing and success rate
Introduce self collision constraint implementation - use bullet etc.
INtroduce the grid map integration
Gradient clipping
Cubic interpolation and open source implementation
\end{comment}
%
%
%%%%%%%%%%%%%%%%%%%%%%%%%%%%%%%%%%%%%%%%%%%%%%%%%%%%%%%%%%%%%%%%%%%%%%%%%%%%%%%%
%
%
%
%
\section{RESULTS}
\label{sec::results}
We tested our planner on \ac{HEAP} \cite{HEAP}, which is a customized Menzi Muck M545 walking excavator; five limbs, 25 joints, and a floating base make it a challenging test bench. The whole planning pipeline is implemented in C++ programming language, and tests are performed on the Intel Xeon E3-1535M processor with 32 GB of RAM. 
%
%
\begin{comment}
The initialization step of our planner leverages OMPL \cite{sucan2012open} planning algorithms implementation. The refinement step is based on Ipopt solver \cite{wachter2006implementation} and our previous work \cite{jelavic2019whole}. All the rigid body calculations are performed using the RBDL library \cite{felis2017rbdl}.
\end{comment}
%
%
\subsubsection{Roadmap Generation} We use a roadmap size of 300 vertices per leg and about 3000 vertices for the arm since it has a much larger workspace (see Fig.~\ref{fig::roadmaps}). More vertices yield better workspace approximation; consequently, finding stable configurations is more likely. However, with more vertices, more computation is required to find stable configurations. We found the proposed number of vertices to be enough for finding solutions for the scenarios tested. For each vertex in the roadmap, we attempt the connection to its ten nearest neighbors (in operational space). If the distance to the nearest neighbor is bigger than $d_{max}$, the connection is rejected. For the legs $d_{max}$=\SI{0.3}{\meter} and for the arm $d_{max}=$\SI{1}{\meter}. \ac{PRM} generation takes about 30 minutes, with more than \SI{99}{\percent} of the computation required for the arm roadmap generation. Looking for connections that are shorter than $d_{max}$ is the most computationally intensive operation.
\subsubsection{Terrains} The proposed planning pipeline can compute plans in various terrains (see Fig.~\ref{fig::terrains}, also see the video attached\footnote{\url{https://youtu.be/B-NHY4xwgwY}}). The planner uses the same set of parameters for all scenarios. Traversing such challenging terrains would not be possible with a purely optimization-based planner and handcrafting a contact schedule for those scenarios would be difficult. We show success rates on complex terrain features in Fig.~\ref{fig::evaluations_success}, by averaging five trials for each difficulty. All terrains except flat ground are shown in Fig.~\ref{fig::terrains}: \emph{rough} (middle left image), \emph{gap} (bottom right), \emph{step} (middle right), \emph{hole} (top right). For terrains \emph{flat}, \emph{rough}, \emph{gap} and \emph{step}, the planner was asked to find a path of about \SI{15}{\meter} in length. For terrain \emph{hole}, the length was about \SI{30}{\meter} such that the planner has to navigate around the hole to get to the other side (see Fig.~\ref{fig::terrains}, top right). The maximal distance parameter in the \ac{RRT} was set to \SI{10}{\meter}. We consider planning successful if both initialization and refinement step find a solution. Times and costs in Fig.~\ref{fig::evaluations_cost} have been obtained by averaging ten successful trials across all terrain difficulties. All the plans are found using the same \ac{RRT} optimizing planner with planning time of \SI{4}{\second}. Fig.~\ref{fig::evaluations_cost} shows that the first stage finds initial solutions quickly and that they are close to the optimal ones. Short planning times suggest that the \emph{initialization step} could be used in a receding horizon fashion.
\subsubsection{Importance of Initialization} To quantify effect of good initialization on the convergence, we use linear interpolation (between start and goal pose) as a baseline strategy. The other strategy is using a whole-body plan from the first stage to initialize all the variables. Linear interpolation uses the contact schedule from the first stage since it cannot compute one alone. The computation times until convergence when initialized with linear interpolation are shown in Fig.~\ref{fig::evaluations_optimization} (red color). The times were obtained by averaging ten successful trials with different initializations over all terrains. The computation times variance is caused by different durations of initialization trajectories from the first stage (longer durations require more computation). A good initialization makes a small difference in continuous terrain (\emph{flat} terrain and \emph{rough} terrain with roughness $\pm0.5$). However, it becomes essential for harder terrains (\emph{hole}, \emph{gap}, \emph{step}) since linear interpolation fails to produce a solution. This result corroborates our hypothesis from the Sec.~\ref{sec::introduction} that \ac{TO} can easily handle non-holonomic and nonlinear constraints from the robot model, whereas terrain constraints and contact schedule discovery present challenges for the optimization. 
Unlike agile quadrupeds in \cite{winkler2018gait}, \ac{HEAP} cannot execute full flight phases, which makes the terrain constraints especially challenging.
\subsubsection{Contact Schedule Discovery} The way of terrain traversing can be influenced by tuning the cost function inside \ac{SBP}. This is illustrated in Fig.~\ref{fig::locomotion_shaping}. \ac{HEAP} is commanded to reach the other side of the gap. Trajectory of the base is shown in green color. In the first scenario (Fig.~\ref{fig::contact_happy}), \ac{HEAP} incurs no cost for lifting the legs off. Upon introducing the stepping penalty, \ac{HEAP} realizes that it can use a small bridge to avoid breaking contact on all legs, Fig.~\ref{fig::contact_averse}. This behavior emerges merely by introducing the stepping penalty and without any other modifications. Such flexibility is made possible by optimizing over contact schedule and full-body poses simultaneously. The resulting contact schedules are shown in Fig.~\ref{fig::contact_happy_schedule} and Fig.~\ref{fig::contact_averse_schedule}.
\section{CONCLUSION AND OUTLOOK}
\label{sec::conclusion_outlook}
We present a combined sampling and \ac{TO} based planner for legged-wheeled robots. The sampling-based stage computes whole body configurations and contact schedule; it is based on \ac{RRT} planner and a roadmap that is pre-computed offline. Compared to existing work, the roadmap is extended to store the mapping between joint angles and \ac{CoM}, which allows for quick stability checks in the presence of heavy limbs. Our \ac{SBP} planner achieves fast planning times and could be used interactively. In the second planning stage, \ac{TO} satisfies all system constraints, such as non-holonomic rolling constraint. We integrate elevation maps into \ac{TO} and demonstrate planning on general map representations. Evaluations of the proposed approach suggest that the main difficulty for \ac{TO} stems from terrain/collision (avoidance) constraints and contact schedule planning, problems that are mitigated using the proposed two-stage approach.

Future developments will verify the approach on real machines. Besides, we plan to investigate using \ac{TO} for tracking plans from the first stage in an \ac{MPC} fashion.

%%%%%%%%%%%%%%%%%%%%%%%%%%%%%%%%%%%%%%%%%%%%%%%%%%%%%%%%%%%%%%%%%%%%%%%%%%%%%%%%
%\section*{APPENDIX}


%%%%%%%%%%%%%%%%%%%%%%%%%%%%%%%%%%%%%%%%%%%%%%%%%%%%%%%%%%%%%%%%%%%%%%%%%%%%%%%%

%\section*{ACKNOWLEDGMENT}

%The preferred spelling of the word �acknowledgment� in America is without an �e� after the �g�. Avoid the stilted expression, �One of us (R. B. G.) thanks . . .�  Instead, try �R. B. G. thanks�. Put sponsor acknowledgments in the unnumbered footnote on the first page.

%%%%%%%%%%%%%%%%%%%%%%%%%%%%%%%%%%%%%%%%%%%%%%%%%%%%%%%%%%%%%%%%%%%%%%%%%%%%%%%%

\bibliographystyle{IEEEtran}
\bibliography{IEEEabrv,biblio}

\end{document}